\pdfoutput=1
\documentclass[11pt]{article}
\usepackage[]{EMNLP2023}
\usepackage{times}
\usepackage{latexsym}
\usepackage[T1]{fontenc}
\usepackage[utf8]{inputenc}
\usepackage{microtype}
\usepackage{inconsolata}
\usepackage{booktabs}
\usepackage{graphicx}
\usepackage{makecell}
\usepackage{multirow}
\usepackage{colortbl}
\usepackage{float}

\title{Multi-Stage Pre-training Enhanced by ChatGPT for Multi-Scenario \\ Multi-Domain Dialogue Summarization}

\author{Weixiao Zhou\textsuperscript{\textnormal{\dag}} \quad
		Gengyao Li\textsuperscript{\textnormal{\ddag}}\textsuperscript{\textnormal{\S}} \quad
		Xianfu Cheng\textsuperscript{\textnormal{\dag}} \quad
		\textbf{Xinnian Liang}\textsuperscript{\textnormal{\dag}} \\
		\textbf{Junnan Zhu}\textsuperscript{\textnormal{\ddag}}\textsuperscript{\textnormal{\S}} \quad
		\textbf{Feifei Zhai}\textsuperscript{\textnormal{\P}} \quad
		\textbf{Zhoujun Li}\textsuperscript{\textnormal{\dag}}\textsuperscript{\textnormal{*}} \vspace{0.05cm} \\
        \textsuperscript{\textnormal{\dag}}State Key Lab of Software Development Environment, Beihang University \\
        \textsuperscript{\textnormal{\ddag}}State Key Lab of Multimodal Artificial Intelligence Systems, Institute of Automation, CAS \\
        \textsuperscript{\textnormal{\S}}School of Artificial Intelligence, University of Chinese Academy of Sciences \\
        \textsuperscript{\textnormal{\P}}Fanyu AI Research, Zhongke Fanyu Technology Co., Ltd \vspace{0.00cm} \\
        {\fontsize{11pt}{0pt}\selectfont \texttt{\{wxzhou,buaacxf,xnliang,lizj\}@buaa.edu.cn}} \quad
        {\fontsize{11pt}{0pt}\selectfont \texttt{junnan.zhu@nlpr.ia.ac.cn}}}

\begin{document}
\maketitle
\begin{abstract}
Dialogue summarization involves a wide range of scenarios and domains. However, existing methods generally only apply to specific scenarios or domains. In this study, we propose a new pre-trained model specifically designed for multi-scenario multi-domain dialogue summarization. It adopts a multi-stage pre-training strategy to reduce the gap between the pre-training objective and fine-tuning objective. Specifically, we first conduct domain-aware pre-training using large-scale multi-scenario multi-domain dialogue data to enhance the adaptability of our pre-trained model. Then, we conduct task-oriented pre-training using large-scale multi-scenario multi-domain "\textit{dialogue-summary}" parallel data annotated by ChatGPT to enhance the dialogue summarization ability of our pre-trained model. Experimental results on three dialogue summarization datasets from different scenarios and domains indicate that our pre-trained model significantly outperforms previous state-of-the-art models in full fine-tuning, zero-shot, and few-shot settings\footnote{Our corpus and enhanced pre-trained models can be found at \url{https://github.com/zhouweixiao/MP4} \par \hspace{0.02cm} \textnormal{*}Corresponding Author}.
\end{abstract}

\section{Introduction}
Dialogue summarization is the task of generating a summary from a dialogue \citep{xu-etal-2022-narrate}. Specifically, open-domain dialogue summarization involves various scenarios (e.g., Online-Chat \citep{gliwa-etal-2019-samsum} and Daily-Life \citep{chen-etal-2021-dialogsum}), while customer service dialogue summarization involves different domains (e.g., Tweet \citep{feigenblat-etal-2021-tweetsumm-dialog} and E-commerce \citep{lin-etal-2022-roles}).

Recently, general-purpose pre-trained models have achieved significant success in dialogue summarization tasks \citep{lewis-etal-2020-bart, pmlr-v119-bao20a, beltagy2020longformer}. Furthermore, several task-specific pre-trained models \citep{pmlr-v119-zhang20ae, Zhong_Liu_Xu_Zhu_Zeng_2022} have further improved dialogue summarization. Existing dialogue summarization pre-trained model \citep{Zhong_Liu_Xu_Zhu_Zeng_2022} achieves good performance on long dialogue summarization. However, it still has the following limitations: (1) It is only pre-trained on dialogue corpora that include two domains (i.e., Interview and TV show), making it difficult to apply to dialogue summarization in a wide range of scenarios and domains. (2) It utilizes a window-based denoising task as the pre-training objective, which presents a significant gap with the fine-tuning objective. Simultaneously, existing state-of-the-art (SOTA) models generally improve dialogue summarization by modeling dialogue interactions \citep{lin-etal-2022-roles, tang-etal-2022-confit}, incorporating extra information (e.g., topics and roles) \citep{wang-etal-2022-guiding, kim-etal-2022-mind}, and rewriting dialogues \citep{xu-etal-2022-narrate, fang-etal-2022-spoken}. Although these methods have some effect, they still have limited applicability to downstream datasets in different scenarios and domains and are often difficult to apply within the current pre-training paradigm due to complex model architectures.

To address the limitations of previous works, in this study, our goal is to propose a task-specific pre-trained model for dialogue summarization, which has extremely small gap between the pre-training objective and the fine-tuning objective, enabling it to excellently adapt to downstream datasets from a wide range of scenarios and domains in full fine-tuning, few-shot, and zero-shot settings.

\begin{figure*}[!t]
  \centering
  \includegraphics[width=0.98\linewidth]{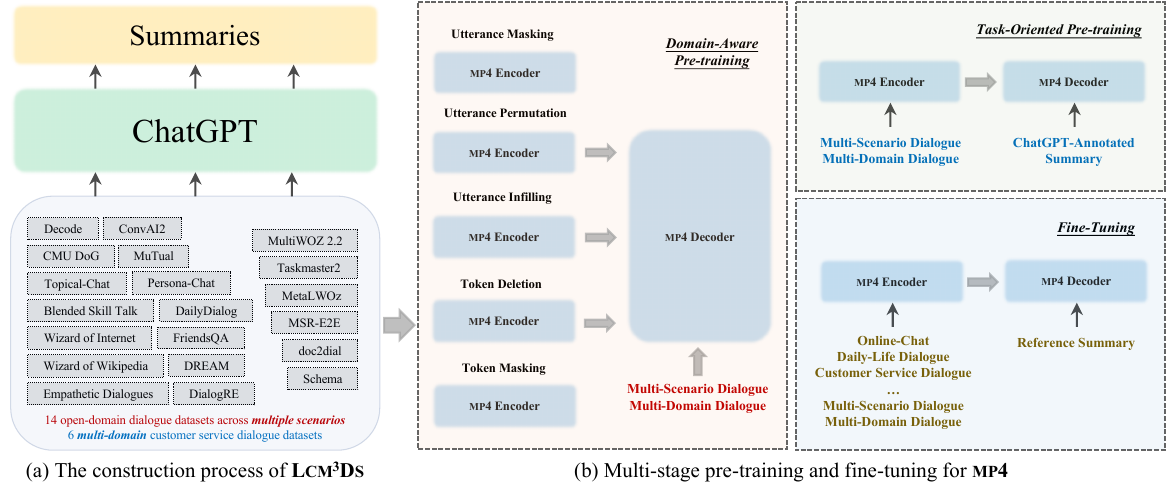}
  \caption{Our pre-training corpus \textbf{\textsc{Lcm}$^\mathbf{3}$\textsc{Ds}} and pre-trained model \textbf{\textsc{mp4}}.}
  \label{fig:1. Introduction Figure 1}
\end{figure*}

Motivated by the above goal, we consider three key components in the implementation of our pre-trained model: model architecture, pre-training corpus, and pre-training strategy. 
For \textbf{model architecture}, our pre-trained model is based on the standard Transformer \citep{NIPS2017_3f5ee243} encoder-decoder architecture and is initialized with BART \citep{lewis-etal-2020-bart}. To capture the underlying role interactions during the dialogue process, we incorporate additional speaker embeddings \citep{gu2020speakeraware, gu-etal-2021-mpc} into token representations. 
For \textbf{pre-training corpus}, we collect 14 open-domain dialogue datasets across multiple scenarios and 6 multi-domain customer service dialogue datasets. Furthermore, due to the development of Large Language Models (LLMs) \citep{zeng2022glm, thoppilan2022lamda, scao2022bloom} and their excellent generative ability, obtaining high-quality "\textit{dialogue-summary}" parallel pre-training data has become possible. Therefore, we utilize ChatGPT \citep{NEURIPS2022_b1efde53} to annotate the collected multi-scenario multi-domain dialogues and obtain corresponding summaries. We refer to our pre-training corpus as \textbf{\textsc{Lcm}$^\mathbf{3}$\textsc{Ds}} (\textbf{L}arge-scale \textbf{C}hatGPT-annotated \textbf{M}ulti-scenario \textbf{M}ulti-domain \textbf{M}ulti-turn \textbf{D}ialogue \textbf{S}ummarization) (see Figure~\ref{fig:1. Introduction Figure 1} (a)). 
For \textbf{pre-training strategy}, we conduct multi-stage pre-training to reduce the gap between the pre-training objective and the fine-tuning objective. Specifically, we first conduct domain-aware pre-training using the dialogue data from \textsc{Lcm}$^3$\textsc{Ds} to enhance the adaptability of pre-trained model to dialogues in multiple scenarios and domains. Then, we utilize the "\textit{dialogue-summary}" parallel data from \textsc{Lcm}$^3$\textsc{Ds} for task-oriented pre-training to enhance the ability of pre-trained model to summarize multi-scenario multi-domain dialogues. We refer to our pre-trained model as \textbf{\textsc{mp4}} (\textbf{M}ulti-stage \textbf{P}re-trained \textbf{M}odel for \textbf{M}ulti-scenario \textbf{M}ulti-domain Dialogue Summarization) (see Figure~\ref{fig:1. Introduction Figure 1} (b)).

We evaluate our pre-trained model on open-domain dialogue summarization datasets from two scenarios (i.e., Online-Chat \citep{gliwa-etal-2019-samsum} and Daily-Life \citep{chen-etal-2021-dialogsum}), as well as a customer service dialogue summarization dataset from a specific domain (i.e., Tweet \citep{feigenblat-etal-2021-tweetsumm-dialog}). The experimental results indicate that \textsc{mp4} significantly outperforms previous SOTA models in full fine-tuning, zero-shot, and few-shot settings, demonstrating remarkable performance improvements.

Our contributions are summarized as follows:
\begin{itemize}
    \item We construct \textsc{Lcm}$^{3}$\textsc{Ds}, which includes a large-scale collection of multi-scenario multi-domain dialogues and their corresponding summaries annotated by ChatGPT.
    \item We propose \textsc{mp4}, a multi-stage pre-trained model for multi-scenario multi-domain dialogue summarization.
    \item Our pre-trained model achieves new state-of-the-art performance on three dialogue summarization datasets from different scenarios and domains in full fine-tuning, zero-shot, and few-shot settings.
\end{itemize}

\begin{table*}[!t]
    \centering
    \small
    \setlength{\tabcolsep}{3.0pt}
    \resizebox{\linewidth}{!}
    {
	\begin{tabular}{l|c|ccc|ccc|ccc|ccc}
	    \toprule[1pt]
	    \multirow{2}{*}{\textbf{Dataset}} & 
	    \multirow{2}{*}{\textbf{Sce./Dom.}} & 
	    \multirow{2}{*}{\textbf{\#Dialogue}} & 
	    \multirow{2}{*}{\thead{\textbf{\#Tokens/} \\ \textbf{dial.}}} & 
	    \multirow{2}{*}{\thead{\textbf{\#Tokens/} \\ \textbf{summ.}}} & 
	    \multirow{2}{*}{\textbf{\#Comp.}} & 
	    \multirow{2}{*}{\textbf{\#Cov.}} & 
	    \multirow{2}{*}{\textbf{\#Dens.}} & 
	    \multicolumn{3}{c|} {\textbf{\textbf{\% of novel $n$-grams}}} &
	    \multicolumn{3}{c} {\textbf{\textbf{\% of redundant $n$-grams}}} \\ 
	    & & & & & & & & \textit{unigram} & \textit{bigram} & \textit{trigram} & \textit{unigram} & \textit{bigram} & \textit{trigram} \\
		\midrule
	    SAMSum & ODDS-Online & 16,368 & 145.1 & 25.3 & 5.99 & 0.71 & \textbf{1.51} & 34.24 & \textbf{78.98} & \textbf{89.54} & \textbf{11.78} & \textbf{1.04} & \textbf{0.20} \\
		\textsc{DialogSum} & ODDS-Daily & 13,460 & 208.9 & 34.5 & 6.48 & \textbf{0.81} & 2.20 & 26.80 & 63.95 & 81.70 & 19.53 & 6.31 & 2.79 \\
		TWEETSUMM & CSDS-Tweet & 1,087 & 328.8 & 39.5 & 7.54 & 0.75 & 2.39 & 32.83 & 73.31 & 83.73 & 16.99 & 1.32 & 0.24 \\
		\rowcolor{brown!30}
		\textsc{Lcm}$^{3}$\textsc{Ds} & \textbf{Multiple} & \textbf{206,768} & 211.3 & 51.9 & \textbf{4.03} & 0.73 & 1.88 & \textbf{34.54} & 74.19 & 86.17 & 21.87 & 3.00 & 0.66 \\
	  \bottomrule[1pt]
  \end{tabular}
  }
  \caption{Comparison between \textsc{Lcm}$^{3}$\textsc{Ds} and existing dialogue summarization datasets. \textit{Sce.} denotes scenario of Open-Domain Dialogue Summarization (ODDS). \textit{Dom.} denotes domain of Customer Service Dialogue Summarization (CSDS). \textit{\#} represents the average value. \textit{dial.} represents dialogue. \textit{Summ.} represents reference summary or ChatGPT-annotated summary. \textit{Comp.} indicates compression ratio. \textit{Cov.} indicates coverage. \textit{Dens.} indicates density.}
  \label{table:2. LCM3DS Corpus Table 1}
\end{table*}

\section{\textsc{Lcm}$^\mathbf{3}$\textsc{Ds} Corpus}
\subsection{Dialogue Preparation}
\paragraph{Dialogue Collection.}
We collect 20 high-quality human-to-human multi-turn dialogue datasets to construct the dialogue part of \textsc{Lcm}$^{3}$\textsc{Ds}, including 14 open-domain dialogue datasets across multiple scenarios \citep{rashkin-etal-2019-towards, nie-etal-2021-like, dinan2019wizard, komeili-etal-2022-internet, Gopalakrishnan2019, zhang-etal-2018-personalizing, smith-etal-2020-put, diana-etal-2018-the, li-etal-2017-dailydialog, cui-etal-2020-mutual, zhou-etal-2018-dataset, yu-etal-2020-dialogue, yang-choi-2019-friendsqa, sun-etal-2019-dream} and 6 multi-domain customer service dialogue datasets \citep{lee2019multi, Rastogi_Zang_Sunkara_Gupta_Khaitan_2020, byrne-etal-2019-taskmaster, zang-etal-2020-multiwoz, li2018microsoft, feng-etal-2020-doc2dial}. In total, there are 105,426 open-domain dialogues containing over 1M utterances and 101,342 customer service dialogues containing over 1.4M utterances. We provide the details of data statistics for all dialogue datasets in Appendix~\ref{sec:data statistics for dialogue datasets}.

\paragraph{Dialogue Pre-Processing and Cleaning.}
We conduct a series of automated data pre-processing and cleaning to further improve the quality of the dialogues. For pre-processing, we perform the following steps: (1) Normalizing punctuations, special characters, and capitalization in each dialogue. (2) Following previous studies \citep{dinan2019wizard, chen-etal-2021-dialogsum}, we preprocess each dialogue into a dual-turn dialogue format by merging consecutive utterances from the same speaker. For cleaning, we perform the following steps: (1) Removing duplicate and highly similar dialogues using the $Jaccard$ text similarity algorithm. (2) Deleting highly similar dialogues \textbf{between the dialogue datasets and the evaluation datasets} using the same algorithm as in (1), \textbf{ensuring that they have no intersection}. (3) Removing dialogues with less than 4 turns or 32 tokens.

\paragraph{Role Adding.}
In order to standardize the different speaker formats of original dialogues across various datasets, we collect a list containing over 4,000 real names. For each dialogue, we randomly selected several real names from the list to assign a role group (e.g., \textit{Danny} and \textit{Alejandra}), where the number and order of real names in each role group corresponds to the speakers in the original dialogue.

\subsection{Annotation}
\paragraph{Prompt Format.}
We follow the previous study of InstructGPT \citep{NEURIPS2022_b1efde53} by inserting the text "\textit{Tl;dr:}" at the end of each dialogue as a prompt and inputting it into ChatGPT\footnote{We use \textit{gpt-3.5-turbo-0301}} (in zero-shot setting) to obtain annotated summaries. We also investigate the performance of three different prompts for dialogue summarization in zero-shot setting, and the details can be found in Appendix~\ref{sec:prompts for dialogue summarization}.

\paragraph{Role-Replaced Data Augmentation.}
In dialogue summarization, there are multiple scenarios and domains involving different roles. To alleviate this problem, we propose a simple yet effective method that can be extended to dialogue summarization involving any role. Specifically, we directly replace the roles in the dialogues and summaries from \textsc{Lcm}$^{3}$\textsc{Ds} to obtain an augmented parallel corpus. In this study, we perform replacements for two common types of roles, including named coreference and customer service. The example we provide can be found in Appendix~\ref{sec:example of data augmentation}.

\subsection{Data Analysis}
We empirically compare \textsc{Lcm}$^{3}$\textsc{Ds} with existing dialogue summarization datasets \citep{gliwa-etal-2019-samsum, chen-etal-2021-dialogsum, feigenblat-etal-2021-tweetsumm-dialog} based on five metrics \citep{grusky-etal-2018-newsroom, fabbri-etal-2021-summeval}: \textit{Compression Ratio}, \textit{Coverage}, \textit{Density}, \textit{Novelty}, and \textit{Redundancy} (see Table~\ref{table:2. LCM3DS Corpus Table 1}).

Compared to existing dialogue summarization datasets, \textsc{Lcm}$^{3}$\textsc{Ds} exhibits lower \textit{Compression Ratio}, moderate \textit{Coverage}, and lower \textit{Density}, indicating that the summaries maintain a high degree of abstraction while covering the important content of the dialogue and retaining more details and information from the dialogue. Additionally, \textsc{Lcm}$^{3}$\textsc{Ds} shows higher \textit{Novelty} and \textit{Redundancy}, which is mainly caused by the lower \textit{Compression Ratio}. Furthermore, unlike existing small-scale, human-annotated, single-scenario, single-domain dialogue summarization datasets, \textsc{Lcm}$^{3}$\textsc{Ds} is large-scale, ChatGPT-annotated, multi-scenario, and multi-domain.

\begin{figure*}[!t]
  \centering
  \includegraphics[width=0.90\linewidth]{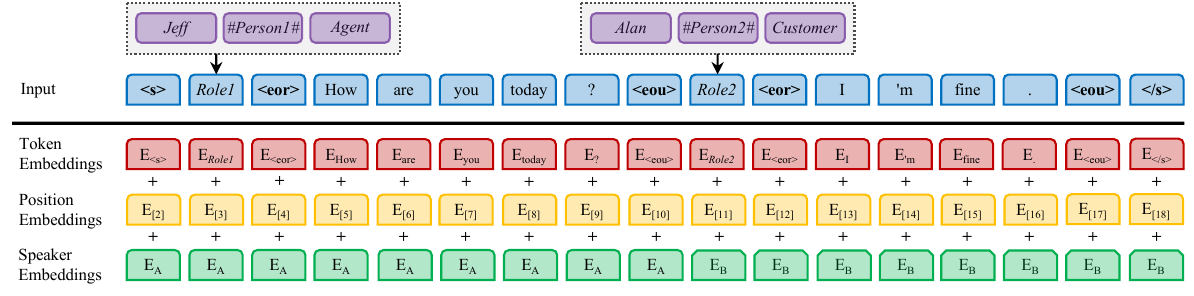}
  \caption{Dialogue modeling of \textsc{mp4}.}
  \label{fig:3. Model Figure 1}
\end{figure*}

\section{Model}
\subsection{Dialogue Modeling}
\textsc{mp4} is based on the standard Transformer \citep{NIPS2017_3f5ee243} encoder-decoder architecture. For multi-turn dialogue modeling, the input embedding of each token is the sum of the corresponding token, position, and speaker embeddings. Figure~\ref{fig:3. Model Figure 1} illustrates the dialogue modeling of \textsc{mp4}.

\paragraph{Input Structure.}
Given the dialogue context $D$, we first concatenate all roles $R_i$ and utterances $U_j$ in the dialogue context with two additional special tokens as a separate, consecutive token sequence:

\noindent $X=\{R_1$<$\mathtt{eor}$>$U_1$<$\mathtt{eou}$>$ \ldots R_m$<$\mathtt{eor}$>$U_n$<$\mathtt{eou}$>$\}$ where the special end-of-role token <$\mathtt{eor}$> and end-of-utterance token <$\mathtt{eou}$> are respectively appended to the end of each role and utterance for multi-turn dialogue separation. Then, we add the start token <$\mathtt{s}$> and the end token <\textit{/}$\mathtt{s}$> around the token sequence $X$ as the input for \textsc{mp4}.

\paragraph{Speaker Embeddings.}
To distinguish utterances in the dialogue context and capture underlying role interactions during the dialogue process, we follow previous dialogue modeling studies \citep{gu2020speakeraware, gu-etal-2021-mpc} and add additional speaker embeddings to token representations. This process is performed alternately based on the role transitions and can be extended to dialogues with an unlimited number of roles. The speaker embeddings are combined with the initial token and position embeddings and then fed into the \textsc{mp4} encoder-decoder framework.

\subsection{Multi-Stage Pre-training}
We conduct multi-stage pre-training to reduce the gap between the pre-training objective and the fine-tuning objective. Domain-aware pre-training aims to enhance the adaptability of \textsc{mp4} to dialogues in multiple scenarios and domains, while task-oriented pre-training aims to enhance the ability of \textsc{mp4} to summarize unstructured spoken multi-scenario multi-domain dialogues into structured written-language summaries.

\subsubsection{Domain-Aware Pre-training}
General-purpose pre-trained models \citep{lewis-etal-2020-bart} are pre-trained on free-form text data with universal pre-training objectives, limiting their ability in specific domains and tasks. Therefore, it is common practice to further train these models with the language modeling objective using text from the target domain to reduce negative impact \citep{zhang-zhao-2021-structural, Whang_Lee_Oh_Lee_Han_Lee_Lee_2021}. In this study, we conduct a domain-aware pre-training stage on \textsc{mp4} using the dialogue data from \textsc{Lcm}$^{3}$\textsc{Ds}. Specifically, we achieve this by modeling a series of \textbf{dialogue reconstruction pre-training objectives} inspired by BART \citep{lewis-etal-2020-bart}.

\paragraph{Token Masking.}
For tokens of each utterance in the dialogue, 20\% of them are randomly sampled and replaced with a special <$\mathtt{mask}$> token.

\paragraph{Token Deletion.}
20\% of the tokens in the dialogue utterances are randomly sampled and deleted.

\paragraph{Utterance Infilling.}
Several utterance spans are randomly sampled, and each span is replaced with a single <$\mathtt{mask}$> token. The length of each utterance span is drawn from the Poisson Distribution ($\lambda = 3$). 0-length spans correspond to the insertion of <$\mathtt{mask}$> tokens.

\paragraph{Utterance Permutation.}
The order of all utterances in the dialogue turns is randomly shuffled. In contrast to previous studies \citep{Zhong_Liu_Xu_Zhu_Zeng_2022, wang-etal-2022-clidsum}, we did not shuffle the order of roles. Therefore, \textsc{mp4} needs to reconstruct the correct order of utterances and ensure the precise alignment between utterances and roles.

\paragraph{Utterance Masking.}
20\% of the utterances in the dialogue are selected and replaced with a special <$\mathtt{uttr}$-$\mathtt{mask}$> token. We did not perform random selection but instead followed the method of PEGASUS \citep{pmlr-v119-zhang20ae} using greedy search to obtain the principal Gap-utterances. During the decoding process, \textsc{mp4} needs to reconstruct the complete dialogue.

\paragraph{Multi-Task Learning.}
The model is trained with a maximum likelihood objective $\mathcal{L}_{\Theta}$. Given the training sample $D = (x, y)$, $\mathcal{L}_{\Theta}$ is defined as
\begin{equation}
    \label{eq:mle}
    \mathcal{L}_{\Theta} = -\sum_{i=1}^{|y|}\log P_{\Theta}(y_i|y_{<i}; x)
\end{equation}
where $\Theta$ is the model parameters, $x$ is the noisy dialogue, and $y$ is the original dialogue.

During each iteration of the multi-task domain-aware pre-training stage, training samples are randomly selected from different pre-training tasks as mini-batches and used to calculate the cumulative loss and optimize the model parameters $\Theta$.

\subsubsection{Task-Oriented Pre-training}
Several task-specific summarization pre-trained models \citep{pmlr-v119-zhang20ae, xiao-etal-2022-primera, Zhong_Liu_Xu_Zhu_Zeng_2022} reduce the gap with downstream datasets by modeling the task-oriented pre-training objective. Specifically, They typically select segments (e.g., gap-sentences or window-based utterances) of the original text (e.g., document or dialogue) as optimization targets for the decoder. Although they have some effects, however, there still exists a significant gap between the segments selected through unsupervised methods and abstractive written-language summaries. In this study, we directly utilize the "\textit{dialogue-summary}" parallel data from \textsc{Lcm}$^{3}$\textsc{Ds} for task-oriented pre-training stage. The learning objective is similar to Eq. (\ref{eq:mle}), where the training sample $D = (x, y)$, with $x$ representing the original dialogue and $y$ representing the summary annotated by ChatGPT.

\section{Experiments}
\subsection{Experimental Setup}
\paragraph{Implementation Details.}
\textsc{mp4} is initialized with BART-large\footnote{\url{https://huggingface.co/facebook/bart-large}} \citep{lewis-etal-2020-bart}. which is a denoising sequence-to-sequence pre-trained Transformer \citep{NIPS2017_3f5ee243} model with 12 layers and 16 attention heads. To facilitate performance comparison, we have implemented four types of \textsc{mp4} models. \textsc{mp4} (\textsc{vanilla}) represents the non-pretrained model, which includes only specialized input structure and speaker embeddings. \textsc{mp4} (\textsc{dap}) denotes domain-aware pre-training applied to \textsc{mp4} (\textsc{vanilla}). \textsc{mp4} (\textsc{top}) signifies task-oriented pre-training applied to \textsc{mp4} (\textsc{vanilla}). \textsc{mp4} (\textsc{dap-top}) indicates multi-stage pre-training applied to \textsc{mp4} (\textsc{vanilla}). More implementation details of pre-training are provided in Appendix~\ref{sec:implementation details of pre-training}.

\paragraph{Downstream Datasets.}
We evaluate the performance of \textsc{mp4} on open-domain dialogue summarization datasets from two scenarios (i.e., Online-Chat and Daily-Life), namely SAMSum \citep{gliwa-etal-2019-samsum} and \textsc{DialogSum} \citep{chen-etal-2021-dialogsum}, as well as a customer service dialogue summarization dataset from a specific domain (i.e., Tweet), namely TWEETSUMM \citep{feigenblat-etal-2021-tweetsumm-dialog}. Table~\ref{table:2. LCM3DS Corpus Table 1} provides the statistics of the downstream datasets. More details are provided in Appendix~\ref{sec:details of downstream datasets}.

\paragraph{Comparison Methods.}
We compare \textsc{mp4} with three types of baselines: extractive models, abstractive models, and previous SOTA models. The following presents the comparison methods.
\begin{itemize}
    \item {\bf Extractive Models.} Based on heuristic algorithms or graph-based algorithms, including \textbf{LONGEST}, \textbf{Lead-3}, and \textbf{TextRank} \citep{mihalcea-tarau-2004-textrank}.
    \item {\bf Abstractive Models.} Based on neural network sequence-to-sequence models, including \textbf{PGNet} \citep{see-etal-2017-get}, \textbf{FastAbs-RL} \citep{chen-bansal-2018-fast}, and \textbf{Transformer} \citep{NIPS2017_3f5ee243}.
    \item {\bf Previous SOTA Models.} State-of-the-art dialogue summarization models based on pre-trained models, including \textbf{BART($\mathcal{D}_\textsc{all}$)} \citep{feng-etal-2021-language}, \textbf{Coref-ATTN} \cite{liu-etal-2021-coreference}, \textbf{BART-ConFiT} \citep{tang-etal-2022-confit}, \textbf{DialSent-PGG} \citep{jia-etal-2022-post}, \textbf{BART-NARR} \citep{xu-etal-2022-narrate}, \textbf{MV-BART} \citep{chen-yang-2020-multi}, \textbf{ReWriteSum} \citep{fang-etal-2022-spoken}, \textbf{BART-SCL} \citep{geng-etal-2022-improving-abstractive}, \textbf{\textsc{UniLMv2}} \citep{pmlr-v119-bao20a}, \textbf{BART} \citep{lewis-etal-2020-bart}, \textbf{LA-BART} \citep{wang2022focused}, and \textbf{BART-MT} \citep{bhattacharjee-etal-2022-multi}.
\end{itemize}

\paragraph{Evaluation Metrics.}
We evaluate the full fine-tuning, zero-shot, and few-shot performance of all models using ROUGE scores\footnote{\url{https://pypi.org/project/py-rouge}} (i.e., R-1, -2, and -L), which are standard evaluation metrics.

\begin{table}[!t]
	\centering
	\small
	\setlength{\tabcolsep}{6.0pt}
	\resizebox{\linewidth}{!}
	{
	\begin{tabular}{lccc}
		\toprule[1pt]
		\textbf{Model} & \textbf{R-1} & \textbf{R-2} & \textbf{R-L} \\
		\midrule
		\multicolumn{4}{c}{\textit{Extractive and Abstractive Models}} \\
		\midrule
		TextRank \citep{mihalcea-tarau-2004-textrank} & 29.27 & 8.02 & 28.78 \\
		PGNet \citep{see-etal-2017-get} & 37.27 & 14.42 & 34.36 \\
		FastAbs-RL \citep{chen-bansal-2018-fast} & 41.03 & 16.93 & 39.05 \\
		Transformer \citep{NIPS2017_3f5ee243} & 42.37 & 18.44 & 39.27 \\
		\midrule
		\multicolumn{4}{c}{\textit{Previous SOTA Models}} \\
		\midrule
		BART($\mathcal{D}_\textsc{all}$) \citep{feng-etal-2021-language} & 53.70 & 28.79 & 50.81 \\
		Coref-ATTN \cite{liu-etal-2021-coreference} & 53.93 & 28.58 & 50.39 \\
		BART-ConFiT \citep{tang-etal-2022-confit} & 53.89 & 28.85 & 49.29 \\
		DialSent-PGG \citep{jia-etal-2022-post} & 53.54 & 28.91 & 50.21 \\
		BART-NARR \citep{xu-etal-2022-narrate} & 53.80 & 28.96 & 50.76 \\
		MV-BART \citep{feng2022survey} & 54.05 & 28.56 & 50.57 \\
		ReWriteSum \citep{fang-etal-2022-spoken} & 54.20 & 27.10 & 50.10 \\
		BART-SCL \citep{geng-etal-2022-improving-abstractive} & 54.22 & 29.87 & 51.35 \\
		\midrule
		\multicolumn{4}{c}{\textit{Our Models}} \\
		\midrule
		BART-large & 53.32 & 28.78 & 50.63 \\
		\textsc{mp4} (\textsc{vanilla}) & 53.41 & 29.23 & 50.97 \\
		\textsc{mp4} (\textsc{dap}) & 53.82 & 29.55 & 51.21 \\
		\textsc{mp4} (\textsc{top}) & 54.56 & 30.33 & 51.70 \\
		\textsc{mp4} (\textsc{dap-top}) & \textbf{54.60} & \textbf{30.57} & \textbf{51.79} \\
		\bottomrule[1pt]
	\end{tabular}
	}
	\caption{Full fine-tuning results on SAMSum test set.}
	\label{table:4. Experiments Table 1}
\end{table}

\subsection{Full Fine-Tuning Evaluation}
To demonstrate the advantages of our pre-trained model with a large amount of training samples, we train the model using the entire training set for full fine-tuning evaluation.

\paragraph{Settings.}
We provide all the hyper-parameters used for fine-tuning and inference in Appendix~\ref{sec:details of fine-tuning and inference}. During the evaluation, for SAMSum, we followed \citep{liu-lapata-2019-text} by testing with the top-3 best checkpoints on the validation set and reporting the average ROUGE scores. For \textsc{DialogSum}, we followed \citep{chen-etal-2021-dialogsum} by reporting the average ROUGE scores between the inference output and multiple reference summaries. For TWEETSUMM, due to limited research and the lack of detailed evaluation procedures in the original paper \citep{feigenblat-etal-2021-tweetsumm-dialog}, we suggest the following evaluation method: (1) During training, use the reference summary with the highest ROUGE-Avg score (i.e., the average value of R-1, R-2, and R-L) between the original dialogue and multiple reference summaries for training. (2) During testing, calculate the average ROUGE scores between the inference output and multiple reference summaries.

\begin{table}[!t]
	\centering
	\small
	\setlength{\tabcolsep}{6.0pt}
	\resizebox{\linewidth}{!}
	{
	\begin{tabular}{lccc}
		\toprule[1pt]
		\textbf{Model} & \textbf{R-1} & \textbf{R-2} & \textbf{R-L} \\
		\midrule
		\multicolumn{4}{c}{\textit{Extractive and Abstractive Models}} \\
		\midrule
		LONGEST & 24.10 & 6.20 & 22.70 \\
		Lead-3 & 27.50 & 6.80 & 27.30 \\
		Transformer \citep{NIPS2017_3f5ee243} & 35.91 & 8.74 & 33.50 \\
		\midrule
		\multicolumn{4}{c}{\textit{Previous SOTA Models}} \\
		\midrule
		\textsc{UniLMv2}-base \citep{pmlr-v119-bao20a} & 47.04 & 21.13 & 45.04 \\
		BART-large \citep{lewis-etal-2020-bart} & 47.28 & 21.18 & 44.83 \\
		BART-NARR \citep{xu-etal-2022-narrate} & 47.52 & 20.82 & 45.10 \\
		LA-BART \citep{wang2022focused} & 47.28 & 21.09 & 45.11 \\
		BART-MT \citep{bhattacharjee-etal-2022-multi} & 47.26 & 21.18 & 45.17 \\
		\midrule
		\multicolumn{4}{c}{\textit{Our Models}} \\
		\midrule
		BART-large & 46.68 & 20.96 & 44.68 \\
		\textsc{mp4} (\textsc{vanilla}) & 46.87 & 21.21 & 44.87 \\
		\textsc{mp4} (\textsc{dap}) & 47.01 & 21.37 & 44.94 \\
		\textsc{mp4} (\textsc{top}) & 47.74 & \textbf{21.84} & 45.77 \\
		\textsc{mp4} (\textsc{dap-top}) & \textbf{48.01} & 21.72 & \textbf{45.92} \\
		\bottomrule[1pt]
	\end{tabular}
	}
	\caption{Full fine-tuning results on \textsc{DialogSum} test set. We report the average of multiple-reference results.}
	\label{table:4. Experiments Table 2}
\end{table}

\begin{table}[!t]
	\centering
	\small
	\setlength{\tabcolsep}{6.0pt}
	\resizebox{\linewidth}{!}
	{
	\begin{tabular}{lccc}
		\toprule[1pt]
		\textbf{Model} & \textbf{R-1} & \textbf{R-2} & \textbf{R-L} \\
		\midrule
		DistilBART \citep{feigenblat-etal-2021-tweetsumm-dialog}* & 37.94 & 19.26 & 33.51 \\
		BART-large (\textit{ours}) & 45.85 & 22.14 & 44.77 \\
		\midrule
		\textsc{mp4} (\textsc{vanilla}) & 45.56 & 22.31 & 44.84 \\
		\textsc{mp4} (\textsc{dap}) & 46.61 & 22.82 & 45.08 \\
		\textsc{mp4} (\textsc{top}) & 46.84 & \textbf{23.63} & 45.74 \\
		\textsc{mp4} (\textsc{dap-top}) & \textbf{46.93} & 23.60 & \textbf{45.82} \\
		\bottomrule[1pt]
	\end{tabular}
	}
	\caption{Full fine-tuning results on TWEETSUMM test set. * denotes results obtained from \citep{feigenblat-etal-2021-tweetsumm-dialog}.}
	\label{table:4. Experiments Table 3}
\end{table}

\paragraph{Results.}
Since most previous SOTA models have not been evaluated on a wide range of dialogue summarization datasets, therefore, we present the full fine-tuning performance on the SAMSum, \textsc{DialogSum}, and TWEETSUMM test sets in Tables~\ref{table:4. Experiments Table 1},~\ref{table:4. Experiments Table 2}, and~\ref{table:4. Experiments Table 3}, respectively. Compared to previous SOTA models, our \textsc{mp4} (\textsc{dap-top}) achieves new state-of-the-art results on all metrics across three downstream datasets from different scenarios and domains, demonstrating significant performance improvements. Specifically, on SAMSum and \textsc{DialogSum}, \textsc{mp4} (\textsc{dap-top}) surpasses the previous SOTA models by improving the R-2 score by \underline{0.70} and \underline{0.54} (29.87→30.57 and 21.18→21.72), respectively. The improvement on TWEETSUMM reaches \underline{1.46} (22.14→23.60). This indicates that multi-stage pre-training can assist the model better adapt to downstream datasets from a wide range of scenarios and domains. Additionally, compared to BART-large, \textsc{mp4} (\textsc{vanilla}) demonstrates stronger performance on most metrics, proving the effectiveness of introducing dialogue modeling. Furthermore, compared to \textsc{mp4} (\textsc{vanilla}), the results of \textsc{mp4} (\textsc{dap}) indicate that domain-aware pre-training can enhance the adaptability of model to dialogues in multiple scenarios and domains. Moreover, \textsc{mp4} (\textsc{top}) achieves significant performance improvements, highlighting the importance of equipping the model with the ability to summarize multi-scenario multi-domain dialogues during the pre-training stage.

\begin{table*}[!t]
    \centering
    \small
    \setlength{\tabcolsep}{5.0pt}
    \resizebox{\linewidth}{!}
    {
	\begin{tabular}{lccccccccc}
	    \toprule[1pt]
	    \multirow{2}{*} {\textbf{Model}} & 
	    \multicolumn{2}{c} {\textbf{SAMSum}} &
	    \multicolumn{2}{c} {\textbf{\textsc{DialogSum}}} & 
	    \multicolumn{2}{c} {\textbf{TWEETSUMM}} \\
	    \cmidrule(r){2-3} %
	    \cmidrule(r){4-5} %
	    \cmidrule(r){6-7} %
	    & \textit{zero-shot} & \textit{few-shot} (10) & \textit{zero-shot} & \textit{few-shot} (10) & \textit{zero-shot} & \textit{few-shot} (10) \\
	    \midrule
	    BART-large (\textit{ours}) & 27.94/8.56/26.28 & 39.47/16.51/38.76 & 25.50/6.10/25.32 & 36.75/12.98/36.04 & 29.42/10.68/28.18 & 43.14/19.46/41.75 \\
		\midrule
		\textsc{mp4} (\textsc{dap}) & 32.48/9.80/31.03 & 41.75/17.34/40.29 & 27.80/6.61/27.42 & 36.89/12.88/36.03 & 31.76/11.30/31.80 & 43.98/20.08/42.31 \\
		\textsc{mp4} (\textsc{dap-top}) & \textbf{42.41}/\textbf{17.15}/\textbf{39.69} & \textbf{48.73}/\textbf{23.63}/\textbf{45.96} & \textbf{38.76}/\textbf{14.59}/\textbf{37.18} & \textbf{40.18}/\textbf{16.22}/\textbf{39.11} & \textbf{38.92}/\textbf{13.16}/\textbf{35.11} & \textbf{45.09}/\textbf{20.20}/\textbf{43.03} \\
	  \bottomrule[1pt]
  \end{tabular}
  }
  \caption{R-1/R-2/R-L results in zero-shot and few-shot settings. For zero-shot setting, we report the results at the optimal summary length limits. For few-shot setting, we report the average results from 5 random runs on 10 training samples (all models share the same seed set).}
  \label{table:4. Experiments Table 4}
\end{table*}

\subsection{Zero- and Few-Shot Evaluation}
Many existing studies that apply pre-trained models to dialogue summarization require a large amount of fine-tuning data, which is often impractical in new scenarios or domains. In contrast, we expect our model to quickly adapt to new scenarios or domains without the need for a large amount of fine-tuning data. To validate this hypothesis, we conduct evaluations in zero-shot (no training samples) and few-shot (10 training samples) settings. Obtaining such a small number of samples is feasible in practice for new scenarios or domains.

\paragraph{Settings.}
We compare the performance of BART-large \citep{lewis-etal-2020-bart}, \textsc{mp4} (\textsc{dap}), and \textsc{mp4} (\textsc{dap-top}) in zero-shot and few-shot settings. In Appendix~\ref{sec:zero- and few-shot evaluation details}, we provide all the hyper-parameters used. Specifically, for zero-shot evaluation, since the models have not been trained on downstream datasets, we report the results of using the optimal summary length limits during inference. For few-shot evaluation, we randomly sample 10 training samples for training. Additionally, to ensure that the results are not affected by sampling variability, we conduct the same experiment five times with different random seeds (shared among all models) and report the average results.

\paragraph{Results.}
The results presented in Table~\ref{table:4. Experiments Table 4} indicate that our pre-trained model achieves significant improvements compared to BART-large. Specifically, for zero-shot results, \textsc{mp4} (\textsc{dap-top}) increases the R-1 score by \underline{14.47} (27.94→42.41), \underline{13.26} (25.50→38.76), and \underline{9.50} (29.42→38.92) on SAMSum, \textsc{DialogSum}, and TWEETSUMM, respectively. Moreover, the zero-shot performance of \textsc{mp4} (\textsc{dap-top}) surpassed the few-shot performance of BART-large on multiple datasets, demonstrating its powerful zero-shot capability. Additionally, the few-shot results also highlight the advantages of \textsc{mp4} (\textsc{dap-top}). indicating that our pre-trained model converges faster than other models even with only 10 training samples.

\begin{table}[!t]
	\centering
	\small
	\setlength{\tabcolsep}{9.0pt}
	\resizebox{\linewidth}{!}
	{
	\begin{tabular}{lccc}
		\toprule[1pt]
		\textbf{Model} & \textbf{R-1} & \textbf{R-2} & \textbf{R-L} \\
		\midrule
		\textsc{mp4} (\textsc{vanilla}) & 53.41 & 29.23 & 50.97 \\
		\quad w/o speaker embeddings & 53.29 & 29.02 & 50.85 \\
		\midrule
		\textsc{mp4} (\textsc{dap}) & 53.82 & 29.55 & 51.21 \\
		\quad w/o \underline{token-level tasks} & 53.74 & 29.25 & 51.05 \\
		\quad w/o utterance infilling & 53.75 & 29.31 & 51.06 \\
		\quad w/o utterance permutation & 53.69 & 29.38 & 51.08 \\
		\quad w/o utterance masking & 53.53 & 29.30 & 50.94 \\
		\midrule
		\textsc{mp4} (\textsc{top}) & 54.56 & 30.33 & 51.70 \\
		\quad w/o CSDS pre-training corpus & 54.47 & 30.12 & 51.69 \\
		\midrule
		\textsc{mp4} (\textsc{dap-top}) & \textbf{54.60} & \textbf{30.57} & \textbf{51.79} \\
		\bottomrule[1pt]
	\end{tabular}
	}
	\caption{Ablation study on SAMSum in full fine-tuning setting. The \underline{token-level tasks} refer to Token Masking and Token Deletion.}
	\label{table:4. Experiments Table 5}
\end{table}

\subsection{Ablation Study}
To further validate the contributions of the fine-grained components in our pre-trained models, we conduct an ablation study on SAMSum in full fine-tuning setting. Table~\ref{table:4. Experiments Table 5} shows the evaluation results.

\paragraph{Speaker Embeddings.}
As the results show, incorporating additional speaker embeddings in dialogue modeling can capture the underlying role interactions during the dialogue process and improve the performance of dialogue summarization.

\paragraph{Domain-Aware Pre-training Objectives.}
As the results show, each domain-aware pre-training objective brings performance improvements. It is worth noting that the utterance masking task has the greatest impact on performance, indicating that completing principal Gap-utterances during dialogue reconstruction is crucial for dialogue summarization.

\paragraph{Impact of CSDS Pre-training Corpus on ODDS.}
We remove the customer service "\textit{dialogue-summary}" parallel data from \textsc{Lcm}$^3$\textsc{Ds} to investigate the impact of this portion of data on the performance of open-domain dialogue summarization. The results show that the model trained without this data exhibits a slight decrease in performance. One possible reason is that a few dialogues in SAMSum also involve customer service topics.

\begin{table}[!t]
	\centering
	\small
	\setlength{\tabcolsep}{8.0pt}
	\resizebox{\linewidth}{!}
	{
	\begin{tabular}{lcccc}
		\toprule[1pt]
		\textbf{Model} & \textbf{Flu.} & \textbf{Conci.} & \textbf{Info.} & \textbf{Comp.} \\
		\midrule
		ChatGPT (\textit{zero-shot}) & \textbf{2.24} & 3.82 & 3.55 & \textbf{1.37} \\
		BART-large (\textit{ours}) & 2.77 & 2.19 & 2.48 & 3.13 \\
		\textsc{mp4} (\textsc{dap-top}) & 2.53 & \textbf{2.02} & \textbf{2.14} & 3.08 \\
		\midrule
		Ground Truth & 2.46 & 1.97 & 1.83 & 2.42 \\
		\bottomrule[1pt]
	\end{tabular}
	}
	\caption{Human evaluation on SAMSum test set.}
	\label{table:4. Experiments Table 6}
\end{table}

\section{Human Evaluation}
We conduct human evaluation to further evaluate the performance of our pre-trained model and strong baselines under various paradigms, as well as the Ground Truth (i.e., \textsc{mp4} (\textsc{dap-top}), ChatGPT, BART-large, Ground Truth). Specifically, we randomly select 50 samples from the test set of SAMSum. Then, we invite 3 participants to rank four candidate summaries according to four metrics: \textit{fluency} (Flu.), \textit{conciseness} (Conci.), \textit{informativeness} (Info.), and \textit{comprehensiveness} (Comp.). The top-ranking indicates the best performance on that metric.

Table~\ref{table:4. Experiments Table 6} shows the results of human evaluation (lower average rank is better). Our pre-trained model outperforms BART-large in all metrics but falls behind the Ground Truth. Specifically, ChatGPT achieves the first rank in \textit{fluency} and \textit{comprehensiveness} for the summaries generated in zero-shot setting, surpassing the Ground Truth. However, it exhibits the weakest performance in \textit{conciseness} and \textit{informativeness}. The main reason for this is that ChatGPT tends to generate longer summaries that describe various aspects of the dialogue, including both important and minor details. Moreover, the longer summaries also contribute to an improved overall impression to some extent.

\section{Related Work}
\paragraph{Dialogue Corpora.}
In general, dialogue data can be obtained from two main sources. One is massive-scale dialogue corpora crawled from web platforms such as Reddit \citep{Baumgartner_Zannettou_Keegan_Squire_Blackburn_2020, henderson-etal-2019-repository} and Twitter \citep{ritter-etal-2010-unsupervised}, which are commonly used for pre-training open-domain chatbots \citep{zhang-etal-2020-dialogpt, adiwardana2020humanlike, roller-etal-2021-recipes, chen-etal-2022-dialogved, henderson-etal-2020-convert}. Another source is a collection of open-source dialogue datasets designed for specific tasks, including open-domain dialogue systems \citep{li-etal-2017-dailydialog, zhou-etal-2018-dataset, diana-etal-2018-the, smith-etal-2020-put, zhang-etal-2018-personalizing, Gopalakrishnan2019, komeili-etal-2022-internet, dinan2019wizard}, task-oriented dialogue systems \citep{lee2019multi, Rastogi_Zang_Sunkara_Gupta_Khaitan_2020, byrne-etal-2019-taskmaster, zang-etal-2020-multiwoz, li2018microsoft, feng-etal-2020-doc2dial}, dialogue comprehension \citep{sun-etal-2019-dream, yang-choi-2019-friendsqa, yu-etal-2020-dialogue, cui-etal-2020-mutual, nie-etal-2021-like, rashkin-etal-2019-towards}, and dialogue summarization \citep{gliwa-etal-2019-samsum, chen-etal-2021-dialogsum, feigenblat-etal-2021-tweetsumm-dialog}.

\paragraph{PTMs for Dialogue Summarization.}
Recently, general-purpose pre-trained models have achieved significant success in dialogue summarization tasks \citep{lewis-etal-2020-bart, 10.5555/3455716.3455856, pmlr-v119-bao20a, beltagy2020longformer}. Furthermore, several task-specific pre-trained models \citep{pmlr-v119-zhang20ae, Zhong_Liu_Xu_Zhu_Zeng_2022} have further improved dialogue summarization. Moreover, existing state-of-the-art dialogue summarization models typically leverage pre-trained models and model the characteristics of dialogues to achieve better results, including modeling dialogue interactions \citep{lin-etal-2022-roles, tang-etal-2022-confit}, incorporating extra information \citep{wang-etal-2022-guiding, kim-etal-2022-mind}, and dialogue rewriting \citep{xu-etal-2022-narrate, fang-etal-2022-spoken}. Although these models are effective, they often have complex model structures that are difficult to apply within the current pre-training paradigm.

\paragraph{Large Language Models.}
More recently, LLMs have attracted widespread attention due to their remarkable performance in various knowledge-intensive NLP tasks \citep{zeng2022glm, NEURIPS2022_b1efde53, thoppilan2022lamda, scao2022bloom}. Through large-scale pre-training on massive text corpora \citep{NEURIPS2020_1457c0d6, wei2022chain}, LLMs possess powerful foundational capabilities. Instruction tuning \citep{10.5555/3455716.3455856, wei2022finetuned, chung2022scaling} helps LLMs in understanding natural language task descriptions. while Reinforcement Learning with Human Feedback (RLHF) \citep{NEURIPS2020_1f89885d, bai2022training} aligns generated text with human preferences.

\section{Conclusion}
In this study, we propose \textsc{mp4}, a multi-stage pre-trained model for multi-scenario multi-domain dialogue summarization. To conduct the pre-training, we construct a large-scale ChatGPT-annotated multi-scenario multi-domain multi-turn dialogue summarization corpus called \textsc{Lcm}$^{3}$\textsc{Ds}. Extensive experimental results demonstrate that \textsc{mp4} exhibits remarkable dialogue summarization capabilities.

\section*{Limitations}
Although we have demonstrated the powerful performance of \textsc{mp4} in multi-scenario multi-domain dialogue summarization, there are still some limitations that provide directions for future work: (1) Due to limitations in computational resources, we did not consider long dialogues when constructing \textsc{Lcm}$^{3}$\textsc{Ds}. Therefore, \textsc{mp4} may be more suitable for short dialogue summarization. (2) \textsc{mp4} is initialized with BART-large, which has only 0.4 billion parameters. In future work, we will consider using larger base models.

\section*{Ethics Statement}
All dialogue data used in this study are sourced from previously published works, and there are no copyright restrictions on their academic use, allowing free online access. Since we used ChatGPT for data annotation, the \textsc{Lcm}$^{3}$\textsc{Ds} corpus we constructed is intended solely for academic research purposes. Additionally, \textsc{mp4} is initialized with the weights from BART-large. Therefore, \textsc{mp4} may exhibit biases and harmful behaviors commonly observed in language models.

\section*{Acknowledgement}
We are grateful to Yu Zhou and Rongping Chang for their valuable contributions to the revision of this paper.

\bibliography{anthology, custom}
\bibliographystyle{acl_natbib}

\clearpage
\appendix
\section{Prompts for Dialogue Summarization}
\label{sec:prompts for dialogue summarization}
We utilize ChatGPT as an unsupervised summarizer to annotate dialogues in the SAMSum \citep{gliwa-etal-2019-samsum} test set using three different prompt formats in zero-shot setting:
\begin{itemize}
    \item {\bf Preceding Prompt.} Insert the prompt "\textit{Summarize the following dialogue into a short summary:}" before the dialogue.
    \item {\bf InstructGPT Prompt.} Following InstructGPT \citep{NEURIPS2022_b1efde53} insert the prompt "\textit{Tl;dr:}" after the dialogue.
    \item {\bf Subsequent Prompt.} Insert the prompt "\textit{Summarize the above dialogue into a short summary:}" after the dialogue.
\end{itemize}
\begin{table}[!h]
	\centering
	\small
	\setlength{\tabcolsep}{9.0pt}
	\resizebox{\linewidth}{!}
	{
	\begin{tabular}{lccc}
		\toprule[1pt]
		\textbf{Prompt} & \textbf{Rouge-1} & \textbf{Rouge-2} & \textbf{Rouge-L} \\
		\midrule
		Preceding & 37.90 & 15.19 & 35.89 \\
		InstructGPT & \textbf{42.17} & \textbf{16.84} & \textbf{39.26} \\
		Subsequent & 40.08 & 15.41 & 37.22 \\
		\bottomrule[1pt]
	\end{tabular}
	}
	\caption{Comparison of dialogue summarization performance for different prompts on SAMSum test set, which contains 819 dialogues.}
	\label{table:prompts}
\end{table}
Table~\ref{table:prompts} shows the evaluation results of three different prompts. The \textbf{InstructGPT Prompt} achieves the best performance.

\section{Details of Downstream Datasets}
\label{sec:details of downstream datasets}
\paragraph{SAMSum.}
A natural messenger dialogue summarization dataset containing dialogues created and written down by linguists fluent in English. The dataset includes various topics such as chit-chats, gossiping about friends, arranging meetings, discussing politics, consulting university assignments with colleagues, etc. The sizes of the training, validation, and test sets are 14,731, 818, and 819 respectively.

\paragraph{\textsc{DialogSum}.}
A large-scale dataset for dialogue summarization, which includes face-to-face spoken dialogues covering a wide range of daily-life topics, including schooling, work, medication, shopping, leisure, and travel, etc. Most dialogues take place between friends, colleagues, and between service providers and customers. The sizes of the training, validation, and test sets are 12,460, 500, and 500 respectively.

\paragraph{TWEETSUMM.}
A customer service dialogue summarization dataset, consisting of reconstructed dialogues extracted from the Kaggle Customer Support On Twitter dataset. The Kaggle dataset is a large-scale dataset based on dialogues between consumers and customer support agents on Twitter.com. It covers a wide range of topics and services provided by various companies, including airlines, retail, gaming, music, etc. The sizes of the training, validation, and test sets are 869, 108, and 110 respectively.

\section{Example of Data Augmentation}
\label{sec:example of data augmentation}
\begin{figure}[!h]
  \centering
  \includegraphics[width=0.45\textwidth]{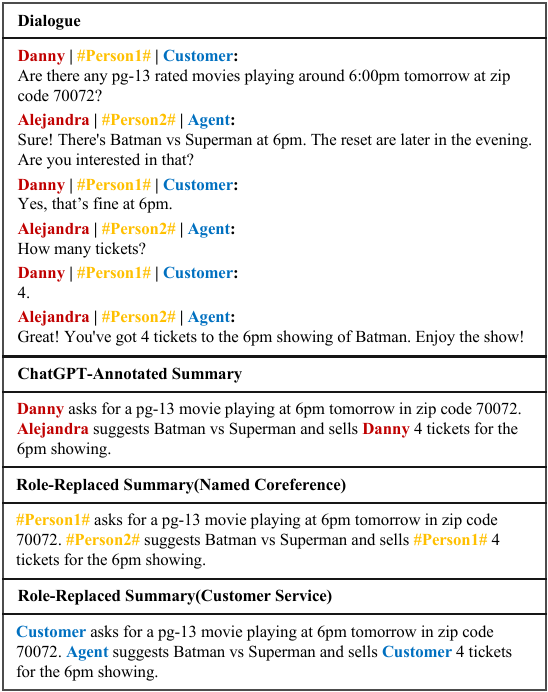}
  \caption{An example of role-replaced dialogue and summary. \textcolor{red}{\textit{Red}} indicates a role group added through role adding, \textcolor{yellow}{\textit{yellow}} indicates named coreference, and \textcolor{blue}{\textit{blue}} indicates roles in customer service.}
  \label{fig:Data Augmentation}
\end{figure}

\section{Implementation Details of Pre-training}
\label{sec:implementation details of pre-training}
\noindent The following are the hyper-parameters used in domain-aware pre-training. \\
\indent $-gpus$ $8$ \\
\indent $-steps$ $5000$ \\
\indent $-batch\_size$ $16$ \\
\indent $-lr$ $3e-05$ \\
\indent $-warmup\_steps$ $500$ \\
\indent $-label\_smoothing$ $0.1$ \\
\indent $-optimizer$ $Adam$ \\
\noindent The following are the hyper-parameters used in task-oriented pre-training. \\
\indent $-gpus$ $8$ \\
\indent $-steps$ $10000$ \\
\indent $-batch\_size$ $16$ \\
\indent $-lr$ $3e-05$ \\
\indent $-warmup\_steps$ $1000$ \\
\indent $-label\_smoothing$ $0.1$ \\
\indent $-optimizer$ $Adam$

\section{Details of Fine-Tuning and Inference}
\label{sec:details of fine-tuning and inference}
\noindent The following are the hyper-parameters used in full fine-tuning setting on SAMSum. \\
\indent $-gpus$ $4$ \\
\indent $-steps$ $1150$ \\
\indent $-batch\_size$ $16$ \\
\indent $-lr$ $3e-05$ \\
\indent $-warmup\_steps$ $100$ \\
\indent $-label\_smoothing$ $0.1$ \\
\indent $-optimizer$ $Adam$ \\
\noindent The following are the hyper-parameters used in full fine-tuning setting on \textsc{DialogSum}. \\
\indent $-gpus$ $4$ \\
\indent $-steps$ $1000$ \\
\indent $-batch\_size$ $16$ \\
\indent $-lr$ $3e-05$ \\
\indent $-warmup\_steps$ $100$ \\
\indent $-label\_smoothing$ $0.1$ \\
\indent $-optimizer$ $Adam$ \\
\noindent The following are the hyper-parameters used in full fine-tuning setting on TWEETSUMM. \\
\indent $-gpus$ $4$ \\
\indent $-steps$ $98$ \\
\indent $-batch\_size$ $16$ \\
\indent $-lr$ $3e-05$ \\
\indent $-warmup\_steps$ $10$ \\
\indent $-label\_smoothing$ $0.1$ \\
\indent $-optimizer$ $Adam$ \\
\noindent All models maintain consistent hyper-parameters across all datasets during inference. \\
\indent $-gpus$ $1$ \\
\indent $-batch\_size$ $32$ \\
\indent $-use\_cache$ $true$ \\
\indent $-max\_length$ $100$ \\
\indent $-min\_length$ $5$ \\
\indent $-beam\_size$ $5$ \\
\indent $-length\_penalty$ $1$ \\
\indent $-no\_repeat\_ngram\_size$ $0$ \\
\indent $-early\_stopping$ $false$

\section{Zero- and Few-Shot Evaluation Details}
\label{sec:zero- and few-shot evaluation details}
\noindent For zero-shot evaluation, the optimal summary length limit hyper-parameter $max\_length$ for SAMSum, \textsc{DialogSum}, and TWEETSUMM is $60$, $40$, and $80$ respectively. Moreover, other hyper-parameters used during inference remain consistent with Appendix~\ref{sec:details of fine-tuning and inference}. For few-shot evaluation (with 10 training samples), we provide the hyper-parameter settings used during training below, while the hyper-parameters used during inference are consistent with Appendix~\ref{sec:details of fine-tuning and inference}. \\
\indent $-gpus$ $1$ \\
\indent $-steps$ $20$ \\
\indent $-batch\_size$ $10$ \\
\indent $-lr$ $3e-05$ \\
\indent $-warmup\_steps$ $0$ \\
\indent $-label\_smoothing$ $0.1$ \\
\indent $-optimizer$ $Adam$ \\
\indent $-seeds$ $3442|3443|3444|3445|3446$

\section{Data Statistics for Dialogue Datasets}
\label{sec:data statistics for dialogue datasets}
Please refer to Table~\ref{table:Data Statistics}.
\begin{table*}[!h]
	\centering
	\small
	\setlength{\tabcolsep}{3.0pt}
	\resizebox{\linewidth}{!}
	{
	\begin{tabular}{lccccccc}
		\toprule[1pt]
		\textbf{Dataset} & \textbf{\#Dialogue} & \textbf{\#Utterance} & \textbf{\#Turns/dial.} & 
		\textbf{\#Tokens/dial.} & \textbf{\#Tokens/turn} & \textbf{\#Tokens/summ.} & \textbf{Sce./\#Dom.} \\
		\midrule
		\multicolumn{8}{c}{\textit{Open-Domain Dialogue}}\\
		\midrule
		Empathetic Dialogues \citep{rashkin-etal-2019-towards} & 22,878 & 98,759 & 4.3 & 86.5 & 20.1 & 34.2 & Empathetic \\
		Decode \citep{nie-etal-2021-like} & 22,119 & 181,580 & 8.2 & 157.6 & 19.2 & 47.3 & Multiple \\
		Wizard of Wikipedia \citep{dinan2019wizard} & 19,287 & 174,624 & 9.1 & 213.7 & 23.5 & 55.8 & Wizard \\
		Wizard of Internet \citep{komeili-etal-2022-internet} & 9,025 & 91,042 & 10.1 & 225.0 & 22.3 & 55.6 & Wizard \\
		Topical-Chat \citep{Gopalakrishnan2019} & 6,524 & 140,544 & 21.5 & 507.7 & 23.6 & 61.2 & Topical \\
		Persona-Chat \citep{zhang-etal-2018-personalizing} & 5,648 & 84,366 & 14.9 & 239.9 & 16.1 & 55.0 & Persona \\
		Blended Skill Talk \citep{smith-etal-2020-put} & 4,662 & 62,581 & 13.4 & 268.1 & 20.0 & 53.8 & Multiple \\
		ConvAI2 \citep{diana-etal-2018-the} & 4,554 & 67,532 & 14.8 & 237.8 & 16.1 & 53.9 & Persona \\
		MuTual \citep{cui-etal-2020-mutual} & 2,719 & 14,134 & 5.2 & 98.7 & 19.0 & 38.1 & Daily \\
		CMU DoG \citep{zhou-etal-2018-dataset} & 2,647 & 58,406 & 22.1 & 422.4 & 19.1 & 67.9 & Movie \\
		DailyDialog \citep{li-etal-2017-dailydialog} & 2,471 & 13,891 & 5.6 & 79.0 & 14.1 & 31.4 & Daily \\
		DialogRE \citep{yu-etal-2020-dialogue} & 1,578 & 20,724 & 13.1 & 249.6 & 19.1 & 49.2 & Chit-Chat \\
		FriendsQA \citep{yang-choi-2019-friendsqa} & 969 & 16,415 & 16.9 & 307.1 & 18.2 & 50.1 & Chit-Chat \\
		DREAM \citep{sun-etal-2019-dream} & 345 & 1,706 & 4.9 & 74.3 & 15.2 & 32.1 & Daily \\
		\midrule
		\multicolumn{8}{c}{\textit{Customer Service Dialogue}} \\
		\midrule
		MetaLWOz \citep{lee2019multi} & 37,860 & 431,580 & 11.4 & 146.0 & 12.8 & 45.2 & 47 \\
		Schema \citep{Rastogi_Zang_Sunkara_Gupta_Khaitan_2020} & 22,559 & 453,320 & 20.1 & 322.7 & 16.1 & 65.1 & 17 \\
		Taskmaster2 \citep{byrne-etal-2019-taskmaster} & 17,083 & 286,930 & 16.8 & 264.7 & 15.8 & 59.8 & 7 \\
		MultiWOZ 2.2 \citep{zang-etal-2020-multiwoz} & 10,330 & 141,162 & 13.7 & 280.4 & 20.5 & 62.6 & 8\\
		MSR-E2E \citep{li2018microsoft} & 8,769 & 68,406 & 7.8 & 150.0 & 19.2 & 46.9 & 3\\
		doc2dial \citep{feng-etal-2020-doc2dial} & 4,741 & 56,221 & 11.9 & 260.7 & 21.9 & 76.9 & 4\\
		\bottomrule[1pt]
	\end{tabular}
	}
	\caption{Data statistics for preprocessed, cleaned and annotated dialogue datasets of \textsc{Lcm}$^{3}$\textsc{Ds}. \textit{\#} denotes average value. \textit{Dial.} represents dialogue. \textit{Summ.} represents ChatGPT-annotated summary. \textit{Sce.} represents scenario of open-domain dialogue. \textit{Dom.} represents domains of customer service dialogue.}
	\label{table:Data Statistics}
\end{table*}

\section{Examples of Generated Summaries}
\label{sec:examples of generated summaries}
\begin{figure*}[!h]
  \centering
  \includegraphics[width=0.95\linewidth]{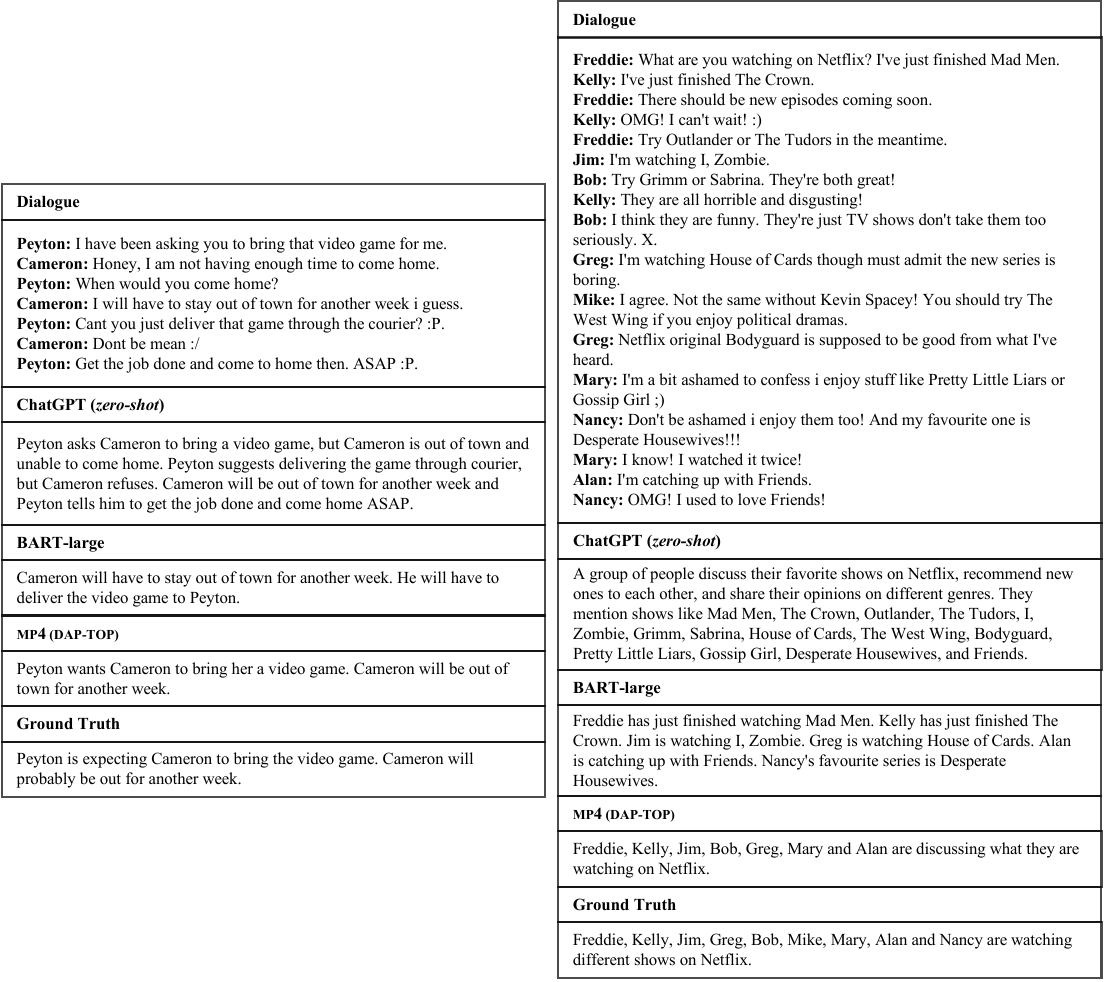}
  \caption{Examples of generated summaries.}
  \label{fig:Case Study}
\end{figure*}

Please refer to Figure~\ref{fig:Case Study}.

\end{document}